\documentclass[journal]{IEEEtran}
\newcommand{\ignore}[1]{}

\let\labelindent\relax

\usepackage[utf8]{inputenc}
\usepackage[T1]{fontenc}

\usepackage{graphicx}
\usepackage{amsmath}
\usepackage{amssymb}
\usepackage{epsfig}
\usepackage{algorithm}  
\usepackage{algpseudocode}
\usepackage{pifont}
\usepackage{multirow,multicol, makecell}
\usepackage{bbm}
\usepackage{mathtools}
\usepackage{diagbox}
\usepackage{enumitem}
\usepackage{soul}
\usepackage{booktabs}
\usepackage{times}
\usepackage{cite}
\usepackage{arydshln}
\usepackage{adjustbox}

\usepackage{xcolor}

\newcommand{\ie}{\textit{i}.\textit{e}.}
\newcommand{\eg}{\textit{e}.\textit{g}.}

\newcommand{\etal}{\textit{et} \textit{al}.}

\usepackage[pagebackref=true,breaklinks=true,colorlinks,bookmarks=false,linkcolor=black,anchorcolor=black,citecolor=black]{hyperref}
\newcommand{\gray}[1]{{\textcolor[RGB]{180,180,180}{#1}}}

\begin{document}
%
\title{GSSF: Generalized Structural Sparse Function for Deep Cross-modal Metric Learning}
\author{Haiwen~Diao,
        Ying~Zhang,
        Shang~Gao,
        Jiawen~Zhu,
        Long~Chen,
        and~Huchuan~Lu, \textit{Fellow}, \textit{IEEE}
\thanks{
Corresponding author: Huchuan Lu. This work is supported by the National Natural Science Foundation of China under grant No. 62293540, 62293542, Liaoning Province Science and Technology Plan No.2023JH26/10200016 and Dalian City Science and Technology Innovation Fund No. 2023JJ11CG001.
H. Diao, S. Gao, J. Zhu, and H. Lu are with the Dalian University of Technology, Dalian, 116024, China.
(Email: diaohw@mail.dlut.edu.cn; gs940601k@gmail.com; jiawen@mail.dlut.edu.cn; lhchuan@dlut.edu.cn).
Y. Zhang is with Tencent Holdings Limited, Shenzhen, 518054, China.
(Email: yinggzhang@tencent.com).
L. Chen is with Hong Kong University of Science and Technology, Hong Kong, 999077, China.
(Email: longchen@ust.hk).
}}
\maketitle
\markboth{IEEE Transactions on Image Processing}{}

\begin{abstract}
Cross-modal metric learning is a prominent research topic that bridges the semantic heterogeneity between vision and language.
Existing methods frequently utilize simple cosine or complex distance metrics to transform the pairwise features into a similarity score, which suffers from an inadequate or inefficient capability for distance measurements.
Consequently, we propose a Generalized Structural Sparse Function to dynamically capture thorough and powerful relationships across modalities for pair-wise similarity learning while remaining concise but efficient.
Specifically, the distance metric delicately encapsulates two formats of diagonal and block-diagonal terms, automatically distinguishing and highlighting the cross-channel relevancy and dependency inside a structured and organized topology.
Hence, it thereby empowers itself to adapt to the optimal matching patterns between the paired features and reaches a sweet spot between model complexity and capability.
Extensive experiments on cross-modal and two extra uni-modal retrieval tasks (\textit{image-text retrieval}, \textit{person re-identification}, \textit{fine-grained image retrieval}) have validated its superiority and flexibility over various popular retrieval frameworks.
More importantly, we further discover that it can be seamlessly incorporated into multiple application scenarios, and demonstrates promising prospects from Attention Mechanism to Knowledge Distillation in a plug-and-play manner.
\end{abstract}
\IEEEpeerreviewmaketitle

\section{Introduction}
\label{sec:introduction}
Recently, tremendous progress in the cross-modal retrieval task aims to learn multi-modal representations and bridge the gap between vision and language, which includes two mainstream architectures. Specifically, some researches~\cite{ITM:SCO,ITM:GXN,ITM:VSRN,ITM:GPO,ITM:WCGL,ITM:AME,ITM:CODER,ITM:MV-VSE,ITM:ESA} employ a dual-encoder architecture with two aggregation modules to encode images and texts independently, and measure their similarities in a joint embedding space. However, their shallow interactions are hampered by recognizing abundant and complicated correspondences across modalities. Consequently, another line of approaches~\cite{ITM:SCAN,ITM:PFAN,ITM:ACME,ITM:CAAN,ITM:MMCA,ITM:DIME,ITM:CMCAN,ITM:NAAF,ITM:RCAR,ITM:SGRAF} concentrates on designing fine-grained interaction modules to model more powerful matching patterns. Note that both of the above sets have a strong need for an accurate, reliable, and efficient distance measurement between pair-wise features across modalities.

\begin{table}[t]
    \centering
    \caption{Evaluations with two representative methods with different metrics on Flickr30K. I $\to$ T indicates R@1 for image-based text retrieval, and vice versa. Time denotes the test time.}
    \label{tab:metrics}
    \setlength{\tabcolsep}{1.4mm}{
    \begin{tabular}{lccccccc}
    \toprule
    \multirow{2}{*}{Metric}
    &{Param.}
    &\multicolumn{3}{c}{VSE~\cite{ITM:VSE++}} 
    &\multicolumn{3}{c}{SCAN~\cite{ITM:SCAN}} \\
    &(M)
    &I\,$\to$\,T &T\,$\to$\,I &Time &I\,$\to$\,T &T\,$\to$\,I &Time\\
    \midrule
    Cosine~\cite{ITM:VSE++} &0 &60.7 &43.5 &4.34s &69.2 &47.5 &75.14s\\
    Sparse~\cite{sparse_metric} &0 &N/A &N/A &N/A &63.4 &43.5 &$\times$4.41\\
    Low-rank~\cite{Low-Rank_metric} &0 &N/A &N/A &N/A &67.7 &45.4 &$\times$2.72\\
    MLP$_{c}$~\cite{ITM:CAMP} &0.5 &-- &-- &$\times$3.16 &-- &-- &$\times$1.37\\
    MLP$_{t}$~\cite{ITM:MTFN} &6.3 &45.5 &29.2 &$\times$3.53 &-- &-- &$\times$2.83\\
    MLP$_{v}$~\cite{ITM:SGRAF} &0.3 &59.0 &\textbf{44.2} &$\times$3.27 &68.1 &50.3 &$\times$1.47\\
    Dense &1.0 &55.5 &40.9 &$\times$1.07 &72.2 &52.1 &$\times$1.05\\
    Dense$_{re}$ &1.0 &57.3 &40.5 &$\times$1.06 &72.4 &52.6 &$\times$1.05\\
    Dense$_{dp}$ &1.0 &56.8 &41.1 &$\times$1.07 &73.7 &54.2 &$\times$1.05\\
    \midrule
    Diag &{\scriptsize $\approx$} 0 &62.1 &44.0 &$\times$1.02 &69.7 &52.4 &$\times$1.01 \\
    B-Diag &{\scriptsize $\leq$} 0.3 &\textbf{62.4} &43.6 &$\times$1.06 &\textbf{77.1} &\textbf{56.7} &$\times$1.04\\
    \bottomrule
    \end{tabular}}
\end{table}

Current cross-modal similarity learning mainly consists of \textit{classification-based} and \textit{embedding-based} measurements. For the former works \cite{ITM:sm-LSTM, ITM:DSPE}, they regard the retrieval as a binary classification problem, and usually adopt Multi-layer Perceptron (MLP) to map the hybrid contents into a logistic activation score. Among them, CAMP~\cite{ITM:CAMP} transforms image-text latent features into a similarity score by a two-layer MLP (MLP$_{c}$), while MTFN~\cite{ITM:MTFN} proposes a multi-semantic tensor fusion network (MLP$_{t}$) to capture the mixed representations from different subspaces. Besides, SGRAF~\cite{ITM:SGRAF} introduces a vector-based similarity function (MLP$_{v}$) to model more detailed associations between different modalities. These complex distance functions derive the correlations of paired features from cross-channel level and polysemous hierarchy, which have shown strong superiority and drawn increasing attention. However, apart from being computationally cumbersome at the retrieval stage, we empirically discover that they suffer from the issues of being sub-optimal and hard to generalize, when applied directly to other works~\cite{ITM:VSE++,ITM:SCAN} in TABLE~\ref{tab:metrics}. In contrast, the embedding-based ones~\cite{ITM:PVSE,ITM:VSRN,ITM:GPO} are simple and universally valid, where existing studies~\cite{ITM:VSE++,ITM:CMPL,ITM:HAL,ITM:AOQ,ITM:MPL} attach more importance to various losses rather than distance metrics that construct a shared space to strengthen representations. Nevertheless, they typically use the widely-used cosine distance and only consider within-channel associations, hindering the capacities of capturing comprehensive dependency and modeling complicated relevancy across channels.
Mahalanobis~\cite{sparse_metric} and low-rank~\cite{Low-Rank_metric} distances consider between-channel variance and feature correlations but require precise estimates across entire data, making it impractical and inaccurate from mini-batches and global calculation after each iteration. They are time-consuming and inaccurate for locally paired features in SCAN due to iterative processing and complex eigen-decomposition.
Hence, we intuitively bring in a fully-parametrized similarity measurement (Dense) as:
\begin{equation}
\label{eq:densescore}
\mathcal{M}_{d}({\boldsymbol{x}}, {\boldsymbol{y}}) = \boldsymbol{x}^{\top}\boldsymbol{W}\boldsymbol{y}\ ,
\end{equation}
where $\boldsymbol{x},\boldsymbol{y}\in\mathbb{R}^{D}$ denote the pairwise features normalized by L2-norm. 
$\boldsymbol{W}\in \mathbb{R}^{D\times D}$ is a learnable matrix and automatically adjusts the relevance degrees between each channel to enhance the matching capability.
Besides, $\boldsymbol{W}$ can be projected into $\boldsymbol{x}$ or $\boldsymbol{y}$ in advance, the Dense metric is also speed-friendly in practice by direct inner product with the pre-computed and stored features during prediction. Since training a dense $\boldsymbol{W}$ becomes intractable in TABLE~\ref{tab:metrics}, we naturally import L2 regularization (Dense$_{re}$)~\cite{OP:adamW} and dropout technology (Dense$_{dp}$)~\cite{Training:Dropout} to implicitly induce sparsity of $\boldsymbol{W}$ that alleviates hard optimization problems~\cite{OP:dropconnect,OP:weight_connection}. While offering moderate advantages, these non-coercive constraints overlook the overlap and dependencies inherent in the data representation~\cite{CP:OrthoReg,CP:SRR-GR,CP:CW-NNK}, resulting in disorganized and unsystematic interactions across channels.

In this paper, we propose to impose the diagonal and block-diagonal constraints on the learned distance matrix $\boldsymbol{W}$, termed as Generalized Structural Sparse Function, which extends the regular one in Eq.~\eqref{eq:densescore} to a structured decorrelation formulation, and generalizes over various application scenarios in a plug-and-play manner.
In particular, the Diagonal Metric (Diag) explicitly capitalizes on the proportional differences in within-channel relevance, facilitating more discriminative distance representations. Conversely, the Block-Diagonal Metric (B-Diag) enhances between-channel connectivity and sparsity by intelligently integrating correspondence within partitions and effectively mitigating associations across partitions. These metrics demonstrate the ability to dynamically adjust to optimal distance metrics for diverse loss functions, delivering remarkable advantages at minimal additional expense.
Our main contributions are summarized as follows:
\begin{itemize}[leftmargin=*,nolistsep]
\item We present a unified embedding-based distance formulation, and thereby introduce an innovative approach called Generalized Structural Sparse Function (GSSF). Within a theoretical elucidation, its Diag and B-Diag items adaptively refine organized and structured interactions across feature channels, and automatically learn the appropriate measurements for paired features over different network frameworks.
\item Our metrics reach a sweet spot between complexity and capability, and remain effective and efficient in a plug-and-play fashion, which can cooperate with existing loss functions and achieve promising improvements over multiple cross-modal and uni-modal datasets at a negligible extra cost.
\item Our metrics can be seamlessly incorporated into a variety of applications, \eg, Attention Mechanism, Fine-grained Alignment, Knowledge Distillation, and Transfer Learning. They consistently surpass the original distance metrics and show encouraging potential for further development.
\end{itemize}

\section{Related Works}
\label{sec:related-works}

\subsection{Deep Metric Learning}
\label{subsec:deep-metric-learning}
Deep metric learning is an algorithm by learning to identify the positive and negative samples and generalize the distributions across domains. Recent researches in uni-modal retrieval mainly focus on objective functions, including sampling selection~\cite{ML:Contrast,ML:Triplet,ReID:Quadruplet,ML:N-pair}, mining negative~\cite{ReID:HardTriplet,ML:Histogram,ML:Triplet,ML:margin-based}, and data proxy~\cite{ML:Proxy-NCA,ML:angular,ML:SoftTriple}.
Particularly, Tian \etal~\cite{Collapse-Aware} proposed collapse-aware triplet decoupling to address weak adversaries and model collapse, while Chechik \etal~\cite{LargeScale} introduced online algorithm for scalable image similarity learning that involves a bilinear similarity measure over sparse representations.
Besides, the embedding-based family~\cite{ML:Triplet,ReID:HardTriplet,ReID:Quadruplet} frequently supervises the relationships within sample tuples via the constraints based on explicit margin~\cite{ML:Contrast,ML:Triplet,ML:margin-based}, softmax function~\cite{ML:N-pair,ML:InfoNCE,CL:SimCLR}
and LogSumExp operation~\cite{ML:Lifted,ML:MS,ML:tuplet,ML:circle}, while the classification-based one~\cite{Datasets:Flickr30k} typically converts representations into logits via learnable weights and optimizes training process by proxy weight matrix~\cite{ML:Proxy-NCA}, hierarchical proxy-based loss~\cite{CloseImitation}, normalized softmax function~\cite{ML:SphereFace,ML:NormFace}, single~\cite{ML:Centerloss,ML:3Dcenter} or multiple~\cite{ML:SoftTriple} per-class centers, multiplicative-angular~\cite{ML:SphereFace}, additive-cosine~\cite{ML:CosineFace1,ML:CosineFace2}, and additive-angular~\cite{ML:angular,ML:ArcFace}
margins. 
Note that some studies import signal-to-noise ratio (SNR) distance~\cite{ML:SNR,ML:PDDM}, relation-aware embedding~\cite{ML:relationalML} and intra-batch connection~\cite{ML:IBC} for better representations during training, while they still utilize Euclidean distance for testing. In contrast, our method selectively amplifies discriminative features by adjusting within-channel variance and inter-channel connections, performing more effectively than their static and unlearnable metrics.

\subsection{Deep Cross-modal Metric Learning}
\label{subsec:deep-cross-modal-metric-learning}
The success of deep metric learning arouses massive studies~\cite{ITM:DSPE,ITM:VSE++,ITM:AOQ,ITM:NCR,ITM:SAM,Dual-stage,CDCM} in cross-modal 2D and 3D fields. One line of research focuses more on designing effective objective functions. For instance, VSE++~\cite{ITM:VSE++} and Poly~\cite{ITM:MPL} adopted online negative mining with maximum and polynomial strategies separately, while CMPL~\cite{ITM:CMPL} and InfoNCE~\cite{VLP:CLIP} employed the cross entropy loss by translating the similarities matrix into bidirectional probability distributions. 
Besides, Yan \etal~\cite{Causality-Invariant} designed causality-invariant interactive mining to capture relationships across samples/modalities in a unified metric space.
In contrast, we concentrate on generalized distance metrics that complement existing loss functions and reveal better matching patterns. Another line of studies gains insights into similarity learning, where~\cite{ITM:VSE++,ITM:PVSE,ITM:CVSE,ITM:GPO,ITM:CODER} employ cosine distance as measurement and achieve decent performance in image-text retrieval. Inspired by the bilinear models~\cite{Bil:MLB,Bil:MFB,Bil:MUTAN,Bil:BAN,Bil:BLOCK}, other approaches~\cite{ITM:CAMP,ITM:MTFN,ITM:SGRAF} develop classification-based metric by the MLP-like module to capture fine-grained connections between pair-wise features. They are less efficient at generalizing across various backbones compared to our metric. Furthermore, we notice that there are some binary distance metrics, including Hamming, Manhattan, Chebyshev, and more universal Minkowski distances in hash retrieval area~\cite{Hash:Review_HM,Hash:SSM,Hash:TripletHash}, where they transform high-dimensional features into compact binary codes and generate similar hash values for similar samples. In contrast, we focus on constructing a concise yet credible metric tailored for high-dimensional data. Our metric acquires a strong capacity of classification-based ones and maintains the simplicity and efficiency characteristic of embedding-based ones.

\begin{figure*}[t]
    \centering
    \begin{tabular}{@{}c}
        \includegraphics[width=0.9\linewidth,trim= 0 372 232 0,clip]{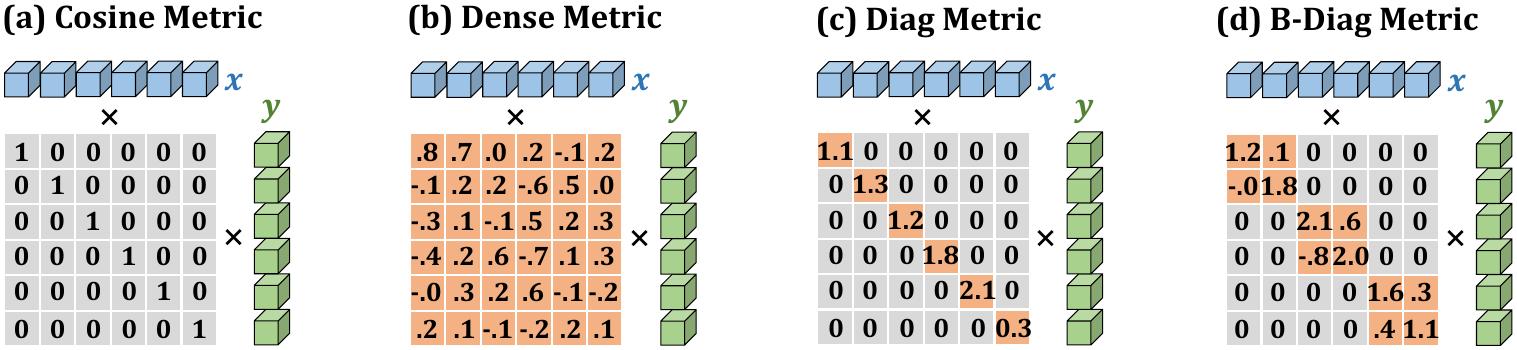} 
    \end{tabular}
    \caption{Illustration of the proposed GSSF metrics. The gray and orange parts indicate the fixed and trainable matrix parameters respectively during the training process. Compared with Cosine and Dense metrics, the Diag item seeks distinct and appropriate weights for within-channel connections, while the B-Diag item further constructs between-channel connectivity and sparsity by structural topology. Both of them are structural sparse strategies of Dense metric.}
	\label{fig:distance_metric}
\end{figure*}

\section{Methodology}
\label{sec:methodology}
Deep Metric learning attempts to learn a distance function where the matched pairs are close together and mismatched ones are far apart.
Sec.\ref{subsec:generalized-metric-learning} describes the detailed formulation of the generalized structural distance function containing two formats of diagonal and block-diagonal items. 
Sec.\ref{subsec:generic-application-scenarios} elaborates several extensions of additional application scenarios.

\subsection{Generalized Metric Learning}
\label{subsec:generalized-metric-learning} 
Recent metric learning pays more emphasis on loss functions~\cite{ITM:VSE++,ITM:MPL,ITM:AOQ,ITM:NCR,ITM:SAM} with cosine score as the distance metric. 
We have validated its insufficiency for fine-grained correspondences across modalities in TABLE~\ref{tab:metrics}. 
Besides, those metrics~\cite{ITM:CAMP,ITM:MTFN,ITM:SGRAF} with millions of parameters occasionally suffer from sub-optimal problems, not to mention their significant time and memory costs. 
Therefore, we introduce innovative Diagonal and Block-Diagonal distance metrics to mine the optimal measurement between paired features, which can be generalized over various application scenarios.

\textbf{Notation.} Given a normalized feature pair ${\boldsymbol{x}}, {\boldsymbol{y}}\in \mathbb{R}^{D}$, structural distance metric extends regular dense one, allowing for learning channel connections inside a constrained space, which constitutes the partition strategy and can be modeled by a binary adjacent matrix $\boldsymbol{U}\in \{0,1\}^{D\times D}$. Such distance metric can be formulated as:
\begin{equation}
\label{eq:structurescore}
\mathcal{M}_{s}({\boldsymbol{x}}, {\boldsymbol{y}}|\boldsymbol{W},\boldsymbol{U}) = \boldsymbol{x}^{\top}(\boldsymbol{W}\odot\boldsymbol{U})\boldsymbol{y}\ ,
\end{equation}
where $\boldsymbol{W}\in \mathbb{R}^{D\times D}$ is a learnable matrix, which weights between-channel value combinations adaptively depending on the distinctions and correlations across feature channels. It is noteworthy that one can flexibly construct diverse distance metrics with Eq.~\eqref{eq:structurescore} by adjusting $\boldsymbol{W}$ and $\boldsymbol{U}$. For example, cosine distance can be formally defined as $\mathcal{M}_{s}({\boldsymbol{x}}, {\boldsymbol{y}}|\boldsymbol{I},\boldsymbol{I})$, where $\boldsymbol{I}$ denotes a fixed identity matrix during training. Drawing from the formula above, we reflect on the critical issues and propose subsequent solutions integral to the development of a comprehensive and efficient distance metric.

\textbf{Discrepancy and Diversity.} With all channels treated equally, the matching results may be interfered by less informative channels. Accordingly, different weight allocation enables to increase within-channel proportion variance and adaptively highlight more discriminative channels, depending on the significance to the optimal distance metric. From this perspective, we develop a new distance metric based on a diagonal matrix (\textbf{Diag}) in Fig.~\ref{fig:distance_metric}(c) as:
\begin{equation}
\label{eq:diagscore}
\begin{split}
    \mathcal{M}_{diag}({\boldsymbol{x}},{\boldsymbol{y}}) &= \mathcal{M}_{s}({\boldsymbol{x}},{\boldsymbol{y}} |\boldsymbol{W},\boldsymbol{I}) \\
    &= \boldsymbol{x}^{\top}(\boldsymbol{W}\odot\boldsymbol{I})\boldsymbol{y}
    = \sum_{m=1}^{D}{w_{m}{x}_{m}{y}_{m}}\ ,
\end{split}
\end{equation}
where ${x}_{m},{y}_{m}$ are the $m$-th channel values of $\boldsymbol{x},\boldsymbol{y}$, and $w_{m}$ indicate the $m$-th diagonal element of $\boldsymbol{W}$. Throughout the training process, the distance metric dynamically adjusts the specialized and differentiated weights for different channels given various databases and frameworks without requiring any manual tuning. It is worth noting that Eq.~\eqref{eq:diagscore} solely addresses relevance values within channels, overlooking any potential connections and transmissions across channels.

\begin{figure*}[t]
    \centering
    \begin{tabular}{@{}c}
        \includegraphics[width=0.9\linewidth,trim= 0 267 45 0,clip]{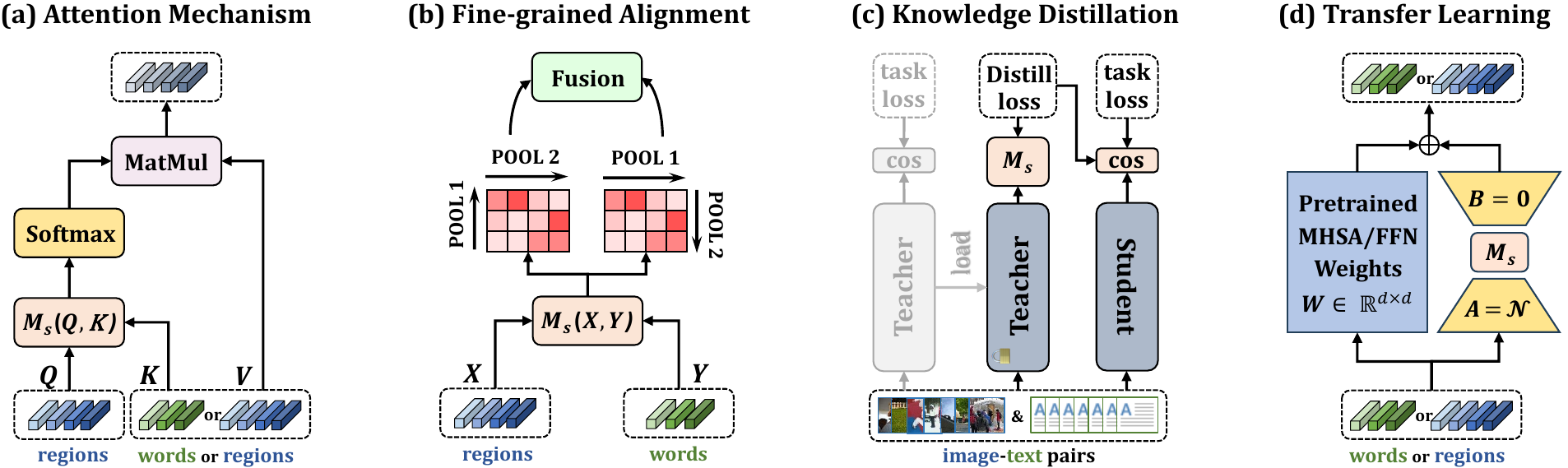} 
    \end{tabular}
   \caption{Illustration of several applications. For (a) Attention Mechanism, it adaptively refines the interactions between query and key from mono- or cross-modality features, while for (b) Fine-grained Alignments, it enhances pairwise similarity scores based on various popular token-wise integration. Besides, we validate its potential by calibrating the teacher's network outputs and adaptation representations during (c) Knowledge Distillation and (d) Transfer Learning.}
   \label{fig:application_scenario}
\end{figure*}

\textbf{Connectivity and Sparsity.}
When it comes to the between-channel dependency, an adjacency matrix is indispensable to bridge the correlations across channels, followed by the risk of a sluggish convergence process and sub-optimal problems from dense connections. Due to the unorganized interaction and less-efficient optimization by implicit sparsity strategies, it is desirable to construct the structured associations that decompose the adjacency matrix, \ie, exploit constrained junctions to decrease redundancy. Hence, we introduce a structural distance metric in a block-diagonal form (\textbf{B-Diag}) in Fig.~\ref{fig:distance_metric}(d) as:
\begin{equation}
\label{eq:b-diagscore}
\begin{split}
    \mathcal{M}_{b\text{-}diag}&({\boldsymbol{x}},{\boldsymbol{y}}) =
    \mathcal{M}_{s}({\boldsymbol{x}},{\boldsymbol{y}}|\boldsymbol{W},\boldsymbol{B})) \\
    &= \boldsymbol{x}^{\top}(\boldsymbol{W}\odot\boldsymbol{B})\boldsymbol{y}
    = \sum_{n=1}^{N}\sum_{i=1}^{d}\sum_{j=1}^{d}{w}_{ij}^{n}{x}_{i}^{n}{y}_{j}^{n}\ .
\end{split}
\end{equation}
Here $\boldsymbol{B} = \text{diag}(\boldsymbol{1}_{1} ,...,\boldsymbol{1}_{N})$, where $\boldsymbol{1}_{n}\in \mathbb{R}^{d\times d}$ is a fixed matrix of ones during training. $N=D/d$ and $d\times d$ are the number and size of each block, while ${w}_{ij}^{n}$ links $i$-th channel of $\boldsymbol{x}^{n}$ and $j$-th channel of $\boldsymbol{y}^{n}$ inside $n$-th block. Besides, $\boldsymbol{W}\odot\boldsymbol{B}$ represents a block-diagonal matrix, dividing all channels into several disjoint partitions and prohibiting the connections across the partitions. With the structured topology, the between-channel information flow is blended in an intrinsic order. The constraints in weight matrices also advocate highly parallel computation and control the trade-off between sparsity and regularity of between-channel connections. \textit{Notably, we initialize $\boldsymbol{W}$ with an identity matrix $\boldsymbol{I}$. In TABLE~\ref{tab:initializtion}, we empirically found that it speeds up convergence and enhances performance by mimicking cosine similarity from the start and escaping optimization issues by random initialization.}

\textbf{Discussion.} Here comes a question: \textit{Why the B-Diag form outperforms the regular one?} Within a mini-batch $\mathcal{D}$, we adopt the $\mathcal{M}_{s}$ in Eq.~\eqref{eq:structurescore} and widely-used hard triplet loss~\cite{ITM:VSE++} for reference. For simplicity, we only analyze the $\mathcal{L}_{\boldsymbol{x}}$ loss on $\boldsymbol{x}\to\boldsymbol{y}$ direction, and vice versa:
\begin{equation}
\label{eq:rankloss}
\mathcal{L}_{\boldsymbol{x}}= \!\!\!\!
\sum_{(\boldsymbol{x},\boldsymbol{y})\in\mathcal{D}}\!\!\! [\ell_{\boldsymbol{x}}]_{+}
=\!\!\!\!
\sum_{(\boldsymbol{x},\boldsymbol{y})\in\mathcal{D}}\!\!\!
[ \gamma  + \boldsymbol{x}^{\top}(\boldsymbol{W}\odot\boldsymbol{U})(\tilde{\boldsymbol{y}} - \boldsymbol{y})]_{+} \ ,
\end{equation}
where $\tilde{\boldsymbol{y}}$ indicates the hardest negative of $\boldsymbol{x}$ in a batch. The gradient with respect to $\boldsymbol{W}$ is derived as:
\begin{equation}
\label{eq:w-grad}
\frac{\partial\mathcal{L}_{\boldsymbol{x}}}{\partial\boldsymbol{W}}
=\sum_{(\boldsymbol{x},\boldsymbol{y})\in\mathcal{D}}
 (\boldsymbol{x}(\tilde{\boldsymbol{y}}-\boldsymbol{y})^{\top}\odot\boldsymbol{U})\cdot\mathbbm{1}(\ell_{\boldsymbol{x}}\geq 0) \ ,
\end{equation}
where if $z$ is true, $\mathbbm{1}(z)=1$, otherwise $\mathbbm{1}(z)=0$. We can discover that $\boldsymbol{U}$ regulates the backward gradient of each element in $\boldsymbol{W}$, and a well-designed one would compulsively refine the cond($\boldsymbol{W}$) to avoid an ill-conditioned solution and reduce the sensitivity of perturbations~\cite{MA:NLA}. Meanwhile, large cut-off connections between channels would inevitably diminish the representation capability of $\boldsymbol{W}$. Hence, the B-Diag form achieves a better trade-off between accuracy and complexity. Notably, our metrics are equivalent to adding an additional vector tensor or fully-connected layer (FC) mapped to $\boldsymbol{x}$ or $\boldsymbol{y}$ in advance, thus preserving the property that the paired distance can be also computed by inner product during the prediction like cosine score at a negligible expense.

\subsection{Generic Application Scenarios}
\label{subsec:generic-application-scenarios}
\textbf{Similarity Prediction.} 
For mono-encoder~\cite{ReID:TransReID,Baseline:Resnet16}, dual-encoder~\cite{ITM:VSE++,ITM:VSRN} or interaction mechanism~\cite{ITM:SCAN,ITM:BFAN,ITM:IMRAM,ITM:NAAF} for retrieval tasks, we simply substitute their original distance with our metrics and reserve the rest architecture for fairness. Moreover, we validate our metrics still exhibit strong adaptability and high versatility under various loss functions including negative mining strategy~\cite{ITM:VSE++,ITM:MPL}, cross-modality projection~\cite{ITM:CMPL} and contrastive training objective~\cite{ML:InfoNCE}.

\textbf{Attention Mechanism.} General attention serves as a transmission between a query and a set of key-value pairs. The query and key vectors mapped by the initial features are utilized to compute the weighted matrix, fusing the corresponding values to enhance the feature representations. Here, we directly insert our metrics into the standard inner-product operation between query and key, and empirically find it beneficial to blend mutual information from different channels inside different blocks, regardless of self-attention~\cite{ITM:VSRN} or cross-attention~\cite{ITM:SCAN,ITM:SGRAF}. Meanwhile, it benefits from being roughly as fast and space-friendly as dot-product operation.

\textbf{Fine-grained Alignment.} In vision-language retrieval~\cite{VLP:FILIP,ITM:TERAN,VLP:WTI}, token-wise integration captures more fine-grained alignments than holistic similarities and performs more efficient inference than interaction frameworks. Existing works start by selecting the maximum values in one row-wise direction and further compress the similarity vector with an averaging, additive, and softly fused operation in another column-wise direction, which corresponds to FILIP~\cite{VLP:FILIP} (Max-Ave), TERAN~\cite{ITM:TERAN} (Max-Sum) and WTI~\cite{VLP:WTI} (Max-Soft) respectively. Here, we reconstruct the alignment matrix based on our distance metrics, which greatly facilitates the thorough and potent cross-modality matching in a plug-and-play fashion.

\textbf{Knowledge Distillation.} Conventional distillation strengthens the capability of the student network by distilling the knowledge flow from a powerful and static teacher network. Here, we only utilize our metrics to calibrate the output of the teacher network during the distillation process, and keep the same similarity metric of the student network with or without distillation supervision. Notably, we combine the task-specific and KL-divergence loss~\cite{KD:DML} with a 1:1 contribution.

\textbf{Transfer Learning.} Fully fine-tuning large networks for downstream tasks is quite costly. Parameter-efficient transfer learning (PETL)~\cite{TL:Adapter-BERT,TL:Prefix-Tuning,TL:LoRA,TL:BitFit,TL:LST,TL:UniPT} offers a solution by tuning partial parameters or inserting a small trainable module into framework during fine-tuning. Here, we adjust the rank decomposition matrices of LoRA~\cite{TL:LoRA} and Adapter~\cite{TL:Adapter-BERT} with our metrics. 

\textbf{Discussion.}
GSSF can generalize across various scenarios with two core design principles:
(1) Adaptive Feature Discrimination: Diag item adapts to varying feature distributions by dynamically adjusting within-channel variances, allowing it to emphasize more discriminative features across diverse scenarios.
(2) Structured Inter-Channel Connectivity: B-Diag item enables a flexible yet structured blending of between-channel information. This is important to adaptively refines channel-wise connections and learn optimal measurements across various network architectures during training process.

\begin{table*}[t]
    \centering
    \caption{Similarity prediction on image-text retrieval benchmark datasets. $\star$ denotes the averaging results of two ensemble models.}
    \label{tab:cocof30k}
    \setlength{\tabcolsep}{2.2mm}{
    \begin{tabular}{lccccccccccccc}
    \toprule
    \multirow{2}{*}{Method}
    &\multicolumn{3}{c}{Flickr30K}
    &\multicolumn{2}{c}{Time}
    &\multicolumn{3}{c}{MSCOCO1K}
    &\multicolumn{3}{c}{MSCOCO5K}
    &\multicolumn{2}{c}{Time}\\
    &I $\to$ T & T $\to$ I &rSum
    &Train &Test
    &I $\to$ T & T $\to$ I &rSum
    &I $\to$ T & T $\to$ I &rSum
    &Train &Test\\
    \midrule
    VSRN\cite{ITM:VSRN}
    &70.2 &53.2 &477.4 & &
    &72.2 &59.9 &507.4
    &48.4 &37.0 &398.0 & &\\
    + Diag
    &70.5 &53.6 &479.2 &+ 2.7\% &+ 2.2\%
    &74.0 &60.6 &\textbf{510.7}
    &51.0 &37.5 &\textbf{405.1} &+ 2.8\% &+ 2.4\%\\
    + B-Diag
    &73.3 &53.3 &\textbf{481.6} &+ 7.3\% &+ 6.1\%
    &73.7 &60.1 &509.9
    &50.6 &37.4 &403.3 &+ 7.5\% &+ 6.8\%\\
    \midrule
    SCAN\cite{ITM:SCAN}$\star$  
    &69.2 &47.5 &464.9 & &
    &71.2 &54.1 &497.3
    &48.2 &34.7 &387.2 & &\\
    + Diag
    &69.7 &52.4 &479.2 &+ 3.5\% &+ 1.3\%
    &77.3 &61.9 &519.0
    &55.4 &39.4 &418.6 &+ 3.7\% &+ 1.3\%\\
    + B-Diag
    &77.1 &56.7 &\textbf{498.1} &+ 8.3\% &+ 4.4\%
    &77.8 &62.1 &\textbf{519.7}
    &56.1 &39.9 &\textbf{421.1} &+ 8.6\% &+ 4.3\%\\
    \midrule
    BFAN\cite{ITM:BFAN}$\star$ 
    &71.9 &53.8 &484.5 & &
    &75.5 &60.0 &513.8
    &54.8 &37.5 &412.8 & &\\
    + Diag
    &75.1 &55.3 &493.0 &+ 3.0\% &+ 1.2\%
    &77.9 &62.5 &521.0
    &57.8 &40.7 &424.4 &+ 3.3\% &+ 1.3\%\\
    + B-Diag
    &77.9 &57.5 &\textbf{499.6} &+ 7.4\% &+ 4.0\%
    &79.5 &62.7 &\textbf{523.3}
    &58.7 &41.1 &\textbf{428.5} &+ 7.7\% &+ 4.1\%\\
    \midrule
    IMRAM\cite{ITM:IMRAM}
    &71.1 &52.0 &477.7 & &
    &77.1 &62.0 &519.0
    &56.1 &40.4 &421.3 & &\\
    + Diag
    &72.7 &53.7 &485.4 &+ 1.8\% &+ 0.8\%
    &78.2 &62.4 &520.6
    &56.9 &40.7 &\textbf{424.1} &+ 2.0\% &+ 0.8\%\\
    + B-Diag
    &73.8 &54.5 &\textbf{486.3} &+ 4.5\% &+ 2.4\%
    &78.5 &62.3 &\textbf{520.8}
    &57.3 &40.8 &423.8 &+ 4.8\% &+ 2.8\%\\
    \midrule
    NAAF\cite{ITM:NAAF}
    &75.3 &56.3 &492.9 & &
    &76.4 &61.2 &516.1
    &55.3 &39.6 &418.0 & &\\
    + Diag
    &76.6 &57.5 &497.2 &+ 3.3\% &+ 1.9\%
    &77.8 &61.6 &\textbf{518.6}
    &56.3 &40.0 &\textbf{421.0} &+ 3.4\% &+ 2.2\%\\
    + B-Diag
    &77.3 &57.5 &\textbf{497.9} &+ 7.8\% &+ 5.8\%
    &78.1 &61.5 &518.4
    &56.5 &39.9 &419.9 &+ 8.0\% &+ 6.3\%\\
    \midrule
    ESA~\cite{ITM:ESA}
    &82.3 &61.2 &514.5 & &
    &79.2 &63.5 &524.9
    &58.0 &41.2 &429.3 & & \\
    + Diag 
    &82.7 &61.5 &515.8 &+ 2.5\% &+ 2.1\%
    &80.1 &63.9 &\textbf{526.7}
    &58.6 &41.6 &\textbf{431.5} &+ 2.7\% &+ 2.2\%\\
    + B-Diag 
    &83.1 &62.0 &\textbf{516.9} &+ 7.0\% &+ 6.0\%
    &79.8 &63.8 &526.4
    &58.5 &41.4 &431.0 &+ 7.3\% &+ 6.0\%\\
    \bottomrule
    \end{tabular}}
\end{table*}

\begin{table*}[t]
    \centering
    \caption{Similarity prediction on conventional and occluded person re-identification, and fine-grained image retrieval datasets.}
    \label{tab:msmt_duke_sop}
    \setlength{\tabcolsep}{1.5mm}{
    \begin{tabular}{lccccccccccccccc}
    \toprule
    \multirow{2}{*}{Method}
    &\multicolumn{3}{c}{MSMT17}
    &\multicolumn{2}{c}{Time}
    &\multicolumn{3}{c}{Occluded-Duke}
    &\multicolumn{2}{c}{Time}
    &\multicolumn{3}{c}{SOP}
    &\multicolumn{2}{c}{Time}\\
    &mAP &R@1 &R@5
    &Train &Test
    &mAP &R@1 &R@5
    &Train &Test
    &R@1 &R@10 &R@10$^{2}$
    &Train &Test\\
    \midrule
    FastReID~\cite{ReID:FastReID}
    &62.62 &81.98 &90.41 & &
    &55.63 &64.12 &79.10 & &
    &79.72 &90.42 &95.69 & &\\
    + Diag
    &62.84 &82.91 &91.17 &+ 3.1\% &+ 1.6\%
    &55.84 &65.25 &79.86 &+ 3.6\% &+ 1.8\%
    &80.68 &91.21 &95.86 &+ 3.4\% &+ 1.6\%\\
    + B-Diag
    &62.95 &82.88 &91.13 &+ 6.7\% &+ 4.3\%
    &55.32 &65.02 &79.46 &+ 7.5\% &+ 4.8\%
    &80.66 &91.30 &95.88 &+ 7.2\% &+ 4.5\%\\
    \bottomrule
    \end{tabular}}
\end{table*}

\section{Experiments}
\label{sec:experiments}
Extensive experiments are conducted to validate the general effectiveness of our metrics.
We insert them into many popular works and validate the potential applications comprehensively.

\subsection{Datasets and Settings}
\label{subsec:datasets-and-settings}

\textbf{Datasets.} We evaluate metrics on image-text retrieval with Flickr30K~\cite{Datasets:Flickr30k} and MSCOCO~\cite{Datasets:MSCOCO}, person re-identification with MSMT17~\cite{Datasets:MSMT17} and Occluded-Duke~\cite{Datasets:Occluded-Duke}, fine-grained image retrieval with Stanford Online Products (SOP)~\cite{ML:Lifted}.

\textbf{Protocols.} We adopt Recall at K (R@K), mAP, and rSum as evaluations. R@K is the percentage of queries whose ground-truth is among the top K candidates, while rSum is the sum of all R@K (K=1, 5, 10) on sentence and image retrieval.

\textbf{Implementation Details.} We re-implement related works with their publicly released codes and keep their original training configurations, including feature dimension, type of optimizer, mini-batch size, learning rate, decaying flow, and hyper-parameter tuning. We conduct all experiments on two GeForce RTX 3090Ti (24G). For holistic feature matching~\cite{ITM:VSRN,ReID:FastReID}, the rate $N$ ($D/d$) is set as 1024 but 4 for the rest of local feature interaction~\cite{ITM:SCAN,ITM:BFAN,ITM:IMRAM,ITM:NAAF} reported in Sec.~\ref{subsec:ablation-study}.

\subsection{Applications and Analysis}
\label{subsec:applications-and-analysis} 

\textbf{Evaluating on Similarity Prediction.}
TABLE~\ref{tab:cocof30k} and~\ref{tab:msmt_duke_sop} investigate 7 popular works~\cite{ITM:VSRN,ITM:SCAN,ITM:BFAN,ITM:IMRAM,ITM:NAAF,ReID:FastReID} with our metrics on five cross-modal and uni-modal datasets. 
\begin{itemize}[leftmargin=*,nolistsep]
    \item \textit{Visual Semantic Reasoning (VSRN)}~\cite{ITM:VSRN}, an embedding-based method that adopts graph reasoning to generate visual-semantic features and final representations sequentially.
    \item \textit{Stacked Cross Attention (SCAN)}~\cite{ITM:SCAN}, an interaction-based method that develops full latent alignments between each image region and text word to infer the similarity score.
    \item \textit{Bidirectional Focal Attention (BFAN)}~\cite{ITM:BFAN}, an interaction-based work that introduces focal and equal attention modules to eliminate unrelated cross-modality correspondence.
    \item \textit{Iterative Matching with Recurrent Attention Memory (IMRAM)}~\cite{ITM:IMRAM}, an interaction-based strategy where image-text relationships are measured with multiple steps of alignments iteratively refined by a memory distillation unit.
    \item \textit{Negative-aware Attention Framework (NAAF)}~\cite{ITM:NAAF}, a light-weight interaction-based method that learns an adaptive boundary, explicitly capturing matched and mismatched fragments, which equally contribute to the final score.
    \item \textit{External Space Attention (ESA)}~\cite{ITM:ESA}, an embedding-based approach that involves an element-wise aggregated attention to obtain adaptive representations across modalities.
    \item \textit{Transformer-based Object Re-Identification (TransReID)} \cite{ReID:TransReID}, an embedding-based framework that reorganizes image patches via shift and shuffle operations, and aggregates patch features by involving several learnable embeddings.
\end{itemize}

\begin{table*}[t]
    \begin{minipage}[t]{0.24\textwidth}
        \centering
        \caption{Attention mechanism with region-region/region-word interactions on Flickr30K.}
        \label{tab:attention} 
        \setlength{\tabcolsep}{0.5mm}{
        \begin{tabular}{lccc}
        \toprule
        {Method}
        &I $\to$ T & T $\to$ I &rSum \\
        \midrule
        VSRN\cite{ITM:VSRN} 
        &70.2 &53.2 &477.4\\
        + Diag
        &70.0 &54.0 &480.6\\
        + B-Diag
        &72.5 &54.2 &\textbf{481.2}\\
        \midrule
        SCAN\cite{ITM:SCAN}$\star$ 
        &69.2 &47.5 &464.9\\
        + Diag
        &69.8 &50.1 &472.9 \\
        + B-Diag
        &74.1 &53.4 &\textbf{485.7}\\
        \midrule
        SGRAF\cite{ITM:SGRAF}$\star$ 
        &77.5 &58.5 &499.3\\
        + Diag
        &78.4 &58.8 &500.9\\
        + B-Diag
        &79.2 &59.6 &\textbf{504.4}\\
        \bottomrule
        \end{tabular}}
    \end{minipage}
    \begin{minipage}[t]{0.24\textwidth}
        \centering
        \caption{Fine-grained alignment with various token-wise similarity aggregations on Flickr30K.}
        \label{tab:alignment}
        \setlength{\tabcolsep}{0.5mm}{
        \begin{tabular}{lccc}
        \toprule
        {Strategy}
        &I $\to$ T & T $\to$ I &rSum \\
        \midrule
        Max-Ave\cite{VLP:FILIP}
        &71.3 &54.6 &486.1\\
        + Diag
        &75.6 &56.7 &497.1\\
        + B-Diag
        &75.7 &57.5 &\textbf{499.4}\\
        \midrule
        Max-Sum\cite{ITM:TERAN}
        &74.0 &51.8 &485.6\\
        + Diag
        &75.2 &54.6 &492.0\\
        + B-Diag
        &77.3 &55.4 &\textbf{495.8}\\
        \midrule
        Max-Soft\cite{VLP:WTI}
        &68.7 &52.6 &477.2\\
        + Diag
        &73.8 &56.5 &496.1\\
        + B-Diag
        &76.3 &57.9 &\textbf{501.0}\\
        \bottomrule
        \end{tabular}}
    \end{minipage}
    \begin{minipage}[t]{0.24\textwidth}
        \centering
        \caption{Knowledge distillation with two embedding/interaction approaches on Flickr30K.}
        \label{tab:distillation}
        \setlength{\tabcolsep}{0.5mm}{\renewcommand{\arraystretch}{0.96}{
        \begin{tabular}{lccc}
        \toprule
        {Method}
        &I $\to$ T & T $\to$ I &rSum \\
        \midrule
        T: VSRN &70.2 &53.2 &477.4\\
        S: VSE &60.7 &43.5 &437.5\\
        + Distill&61.7 &44.8 &441.9\\
        ++ Diag &64.3 &44.9 &\textbf{445.0}\\
        ++ B-Diag &62.1 &44.8 &443.9 \\
        \midrule
        T: NAAF &75.3 &56.3 &492.9\\
        S: SCAN$_{t2i}$ &66.7 &47.9 &460.3\\
        + Distill &65.1 &47.4 &459.6\\
        ++ Diag &67.2 &48.7 &463.9\\
        ++ B-Diag &68.0 &48.3 &\textbf{464.1}\\
        \bottomrule
        \end{tabular}}}
    \end{minipage}
    \begin{minipage}[t]{0.24\textwidth}
        \centering
        \caption{Adaptation to multiple loss functions via the ensemble models SCAN$\star$ on Flickr30K.}\label{tab:lossfunction}
        \setlength{\tabcolsep}{0.5mm}{
        \begin{tabular}{lccc}
        \toprule
        {Loss}
        &I $\to$ T & T $\to$ I &rSum\\
        \midrule
        Poly\cite{ITM:MPL}
        &68.1 &48.3 &463.9\\
        + Diag
        &72.3 &51.1 &477.0\\
        + B-Diag
        &74.3 &55.3 &\textbf{490.5}\\
        \midrule
        CMPM\cite{ITM:CMPL}
        &62.1 &41.6 &434.9\\
        + Diag
        &69.9 &50.1 &473.1\\
        + B-Diag
        &72.5 &53.8 &\textbf{485.1}\\
        \midrule
        InfoNCE\cite{VLP:CLIP} 
        &65.7 &48.2 &464.4\\
        + Diag
        &71.0 &53.9 &483.3\\
        + B-Diag
        &72.4 &54.2 &\textbf{486.4}\\
        \bottomrule
        \end{tabular}}
    \end{minipage}
\end{table*}

TABLE~\ref{tab:cocof30k} demonstrates that in cross-modality domains, our distance metrics effectively improve the evaluation metrics in both Diag and B-Diag manners. Specifically, the Diag item can obtain maximum R@1 increments at sentence/image retrieval with 3.2/4.9\% on Flickr30K, 6.1/7.8\% on MSCOCO1K, and 7.2/4.7\% on MSCOCO5K, while the B-Diag item achieves R@1 benefits of at most 7.9/9.2\%, 6.6/8.0\% and 7.9/5.2\% respectively. 
In TABLE~\ref{tab:msmt_duke_sop}, we further report these two metrics on MSMT17, Occluded-Duke, and SOP, which obtain notable performance gains over large and diverse uni-modal benchmarks. 
The valid and consistent improvements validate the general effectiveness and robustness of various state-of-the-art methods on both small and large-scale datasets. 

\textbf{Evaluating on Attention Mechanism.}
TABLE~\ref{tab:attention} shows plug-and-play capabilities on QKV-based attentions including self-attention~\cite{ITM:VSRN} and cross-attention~\cite{ITM:SCAN,ITM:SGRAF}. We discover that both Diag and B-Diag metrics can effectively promote the correlation and interaction between instance features. To take SCAN as an example, the Diag score empowers stronger cross-modality correspondence and improves rSum by 8.0\%, while the B-Diag score further promotes rSum by 21.0\% increase.

\textbf{Evaluating on Fine-grained Alignment.}
TABLE~\ref{tab:alignment} investigates the applications over various token-wise alignments. Our metrics are well compatible with all three operations, exceeding cosine similarity by at least 6.4\% rSum gains. Besides, Max-Soft mode using the B-Diag score produces the best empirical results, marginally suppressing other setups and exhibiting superior matching patterns across modalities.

\textbf{Evaluating on Knowledge Distillation.}
In TABLE~\ref{tab:distillation}, we conduct the distillation between dual-dual (VSRN-VSE) and cross-cross (NAAF-SCAN) methods. We surprisingly discover that even if the teacher network was previously trained via cosine metric, message flow can be optimized by our measurements to produce better distillation guidance. Besides, SCAN struggles to learn the matching pattern of NAAF, which could be because the latter imports an extra negative score mining, making the former hard to imitate and causing noise interference to the distillation process. In contrast, ours can alleviate this issue and obtain a positive distillation gain.

\textbf{Evaluating on Training Objective.}
TABLE~\ref{tab:lossfunction} reports the compatibility with three popular loss functions whose original hyper-parameters are kept without manual adjustments. 
\begin{itemize}[leftmargin=*,nolistsep]
    \item \textit{Max Polynomial loss (Poly)}~\cite{ITM:MPL}, a general weighting loss that assigns larger weight values to harder samples and highlights informative pairs from redundant pairs.
    \item \textit{Cross-modal Projection Matching loss (CMPM)}~\cite{ITM:CMPL}, a projection loss that minimizes KL divergence between matching projection and normalized label distributions.
    \item \textit{Info Noise Contrastive Estimation loss (InfoNCE)}~\cite{VLP:CLIP}, a contrastive loss in a log-softmax form that attempts to classify the positive pair from multiple negative ones.
\end{itemize}
The constant R@1 gains of at least 6.2/6.0\% at sentence and image retrieval based on SCAN~\cite{ITM:SCAN} confirm the powerful applicability of our metrics with various objective functions.

\begin{table}[t]
    \centering
    \caption{Parameter-efficient fine-tuning with VSE$\infty$ on Flickr30K.}
    \label{tab:transfer}
    \setlength{\tabcolsep}{1.8mm}{
    \begin{tabular}{lccccc}
    \toprule
    \multirowcell{2}{Method} 
    & Params. 
    & Memory
    &\multicolumn{3}{c}{Region + BERT} \\
    \cmidrule{4-6}
    &(M)
    &(G)
    &I $\to$ T &T $\to$ I &rSum \\
    \midrule
    \gray{Fully-Tuning}
    &\gray{109.5} &\gray{9.9} 
    &\gray{79.7} &\gray{62.1} &\gray{513.5}
    \\
    \gray{Partially-Tuning} 
    &\gray{0.8} &\gray{1.0}
    &\gray{74.8} &\gray{53.7} &\gray{485.5}
    \\
    \midrule
    LST~\cite{TL:LST}
    &7.5 &4.6
    &77.9 &57.3 &501.9
    \\
    BitFit~\cite{TL:BitFit} 
    &0.9 &8.6
    &77.3 &57.8 &503.9
    \\  
    Prompt~\cite{TL:Prefix-Tuning}
    &10.7 &9.4
    &78.7 &59.0 &508.5
    \\
    FacT~\cite{TL:FacT}
    &0.6 &8.7
    &79.2 &59.3 &508.8
    \\
    AdaLoRA~\cite{TL:AdaLoRA}
    &1.0 &8.8
    &79.8 &60.1 &510.3
    \\
    UniPT~\cite{TL:UniPT}
    &5.9 &3.1
    &80.2 &59.8 &510.5
    \\
    SSF~\cite{TL:SSF}
    &0.2 &8.4
    &80.0 &60.4 &512.8
    \\
    \midrule
    LoRA~\cite{TL:LoRA}
    &1.1 &8.8
    &78.8 &59.6 &508.2
    \\
    + Diag
    &1.1 &8.8
    &79.4 &59.6 &509.7
    \\
    + B-Diag
    &1.3 &8.9
    &79.8 & 59.9 &\textbf{511.2}
    \\
    \hdashline
    Adapter~\cite{TL:Adapter-BERT}
    &2.6 &8.8
    &79.1 &60.5 &511.3
    \\  
    + Diag
    &2.6 &8.8
    &79.6 &60.3 &511.8
    \\
    + B-Diag
    &3.1 &9.0
    &80.2 &60.5 &\textbf{513.1}
    \\
    \bottomrule
    \end{tabular}
    }
\end{table}

\textbf{Evaluating on Transfer Learning.} In TABLE~\ref{tab:transfer}, we apply our transmission to parameter-efficient fine-tuning based on VSE$\infty$ using 36 region features extracted by the offline detector~\cite{Detection:FasterR-CNN} and word features computed by BERT~\cite{TransF:BERT}. During training, BitFit~\cite{TL:BitFit} updates all the bias items of pre-trained layers, while Prompt~\cite{TL:Prefix-Tuning} concatenates 30 randomly-initialized tokens with the original input. Besides, LST~\cite{TL:LST} and UniPT~\cite{TL:UniPT} build a parallel side network detached from the pre-trained backbone for adaptation. Note that LoRA~\cite{TL:LoRA} refines each QKV projection of multi-head attention (MHSA) module, while Adapter~\cite{TL:Adapter-BERT} inserts lightweight blocks after each MHSA and feedforward layer. All the parameters of the pre-trained network are frozen during fine-tuning. We observe that our proposed metrics promote the adaptation process of existing popular strategies, and improve LoRA and Adapter by at most 3.0\% and 1.8\% rSum gains, respectively.

\begin{table*}[t]
\caption{Influence of cosine initialization for different distance metrics based on the ensemble models SCAN$\star$ on Flickr30K. The left two figures, denoted as (a) and (b), depict the convergence of the network during training with $\boldsymbol{W}$ initialization using the identity matrix $\boldsymbol{I}$ or not. The right table demonstrates the final evaluation results corresponding to these configurations.}
\centering
\label{tab:initializtion}
    \adjustbox{valign=t}{
    \begin{minipage}{0.66\textwidth}
    \includegraphics[height=0.25\textwidth,trim= 0 310 55 0,clip]{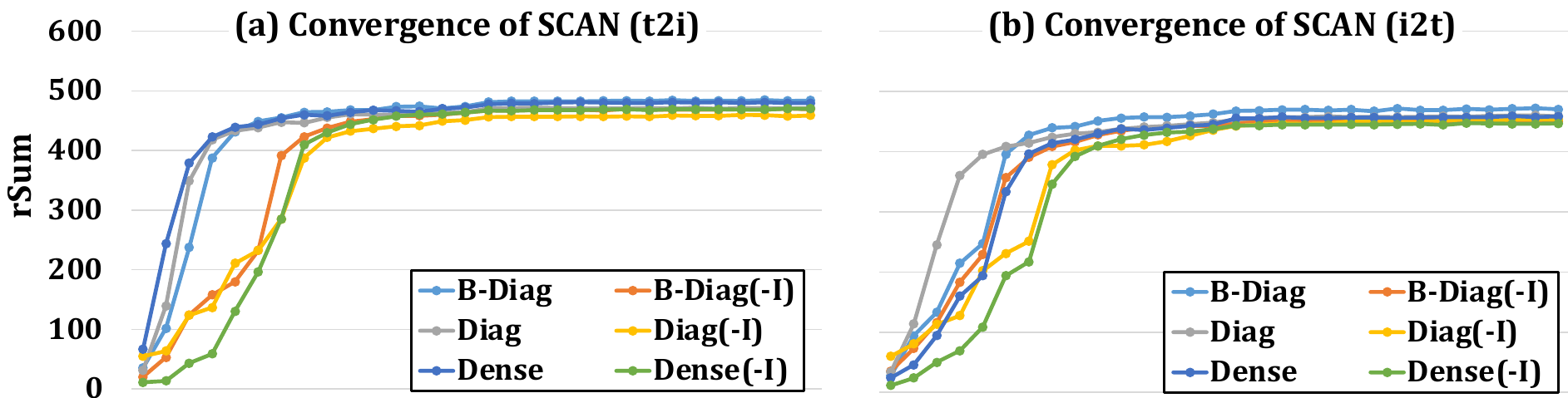}
    \end{minipage}}
    \adjustbox{valign=t}{
    \begin{minipage}{0.3\textwidth}
        \begin{tabular}{lccc}
        \toprule
        {Metric}
        &I $\to$ T & T $\to$ I &rSum\\
        \midrule
        \gray{Cosine}  
        &\gray{69.2} &\gray{47.5} &\gray{464.9} \\
        Dense
        &74.4 &54.5 &490.7\\
        - initial $\boldsymbol{I}$
        &72.2 &52.1 &480.0\\
        Diag
        &69.7 &52.4 &479.2\\
        - initial $\boldsymbol{I}$
        &70.1 &50.6 &473.8\\
        B-Diag
        &77.1 &56.7 &498.1\\
        - initial $\boldsymbol{I}$
        &74.5 &54.8 &488.6\\
        \bottomrule
        \end{tabular} 
    \end{minipage}}
\end{table*}

\begin{figure*}[t]
    \adjustbox{valign=t}{
    \begin{minipage}{0.47\textwidth}
        \begin{tabular}{@{}c}
        \includegraphics[width=\linewidth,trim= 0 30 100 0,clip]{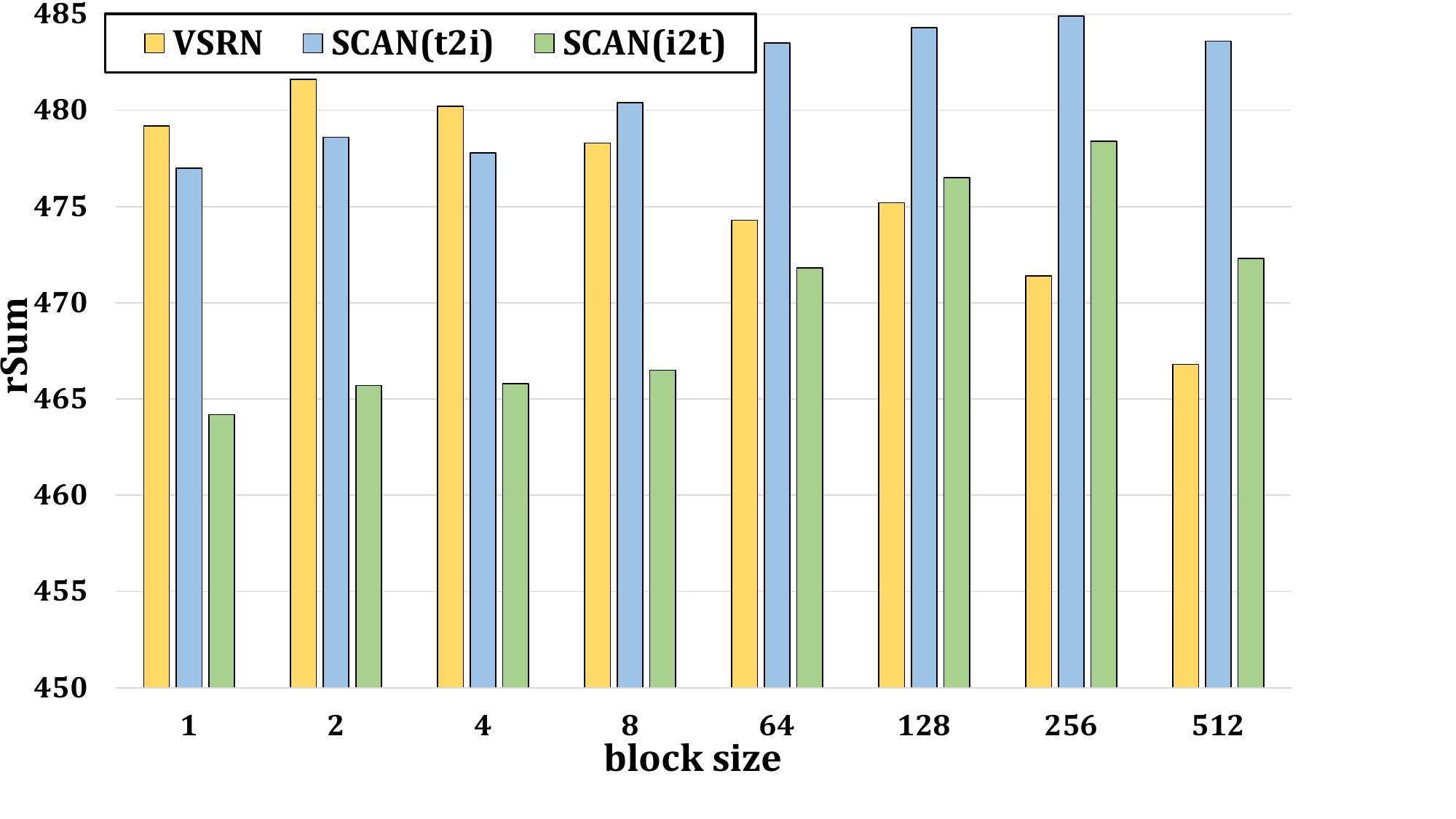}
        \end{tabular}
        \caption{Configuration of various block sizes $d$. The embedding-based VSRN and interaction-based SCAN are utilized to analyze the influence of the ratio $N$ ($D/d$) between the feature dimension $D$ and block size $d$. The smaller the block size $d$, the sparser the channel-wise connection $\boldsymbol{W}$.}
        \label{fig:exp-block_size}
    \end{minipage}}
    \adjustbox{valign=t}{
    \begin{minipage}{0.47\textwidth}
        \begin{tabular}{@{}c}
        \includegraphics[width=\linewidth,trim= 0 273 35 0,clip]{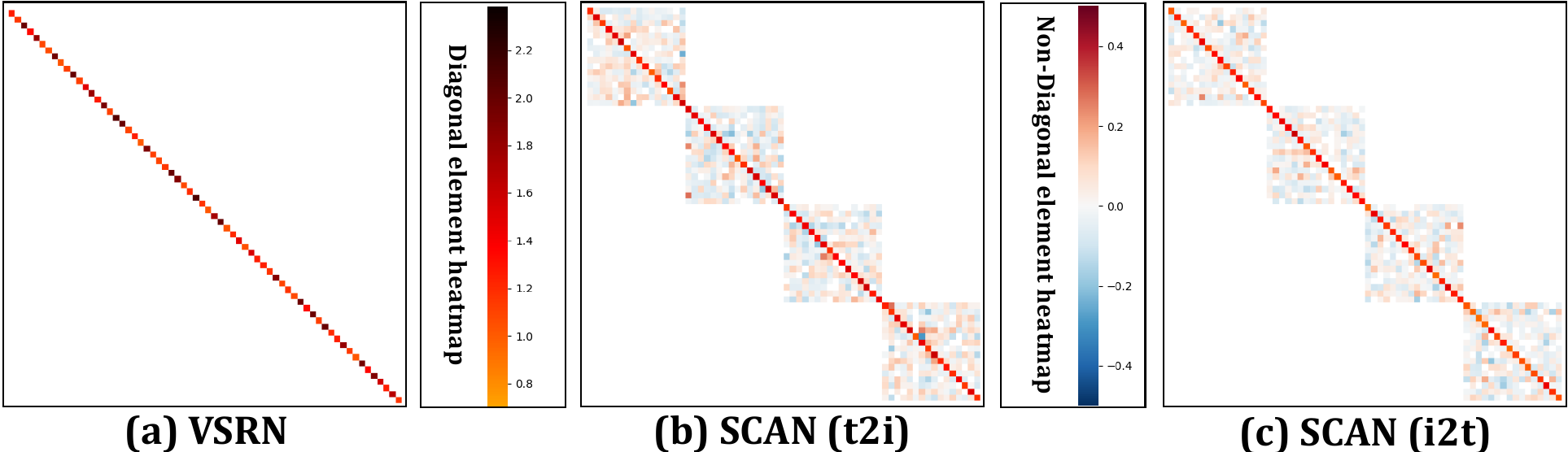} \\
        \includegraphics[width=\linewidth,trim= 0 273 35 0,clip]{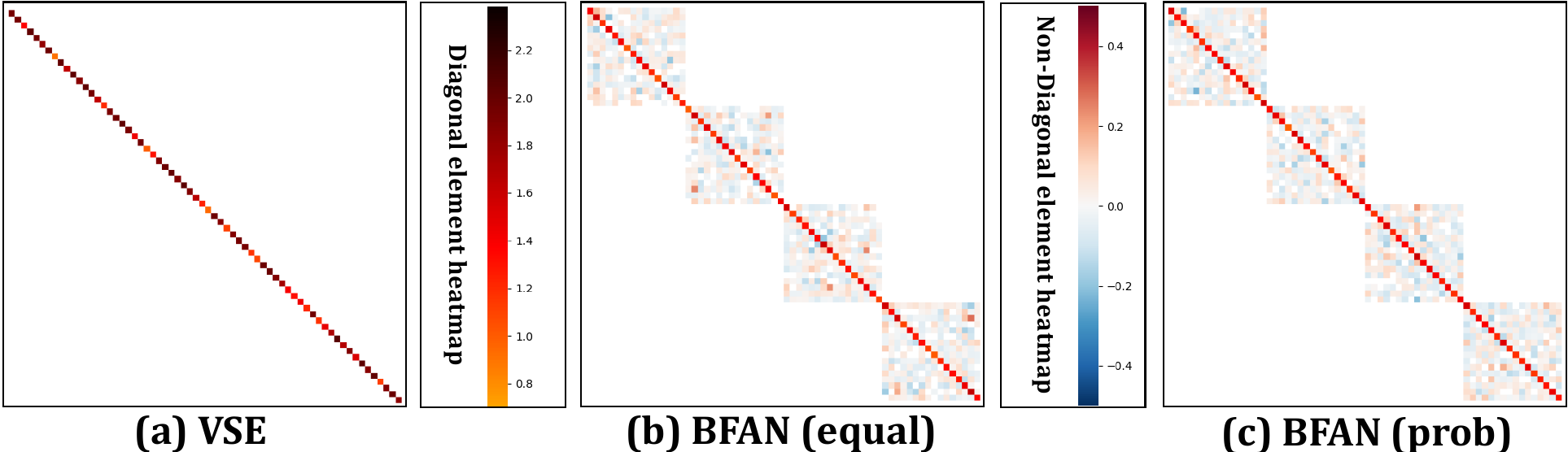} 
        \end{tabular}
        \caption{Visualization of cross-channel weight coefficients learned by the B-Diag metric. For clear visualization, we reduce the size of the matrix $\boldsymbol{W}$ to 64x64 at regular intervals. Note that red and blue zones indicate positive and negative correlations and relationships across channels, respectively.}
        \label{fig:exp-block_weight}
    \end{minipage}}
\end{figure*}

\begin{figure*}[t]
    \centering
    \begin{tabular}{@{}c}
    \includegraphics[width=0.95\linewidth,trim= 0 60 140 0,clip]{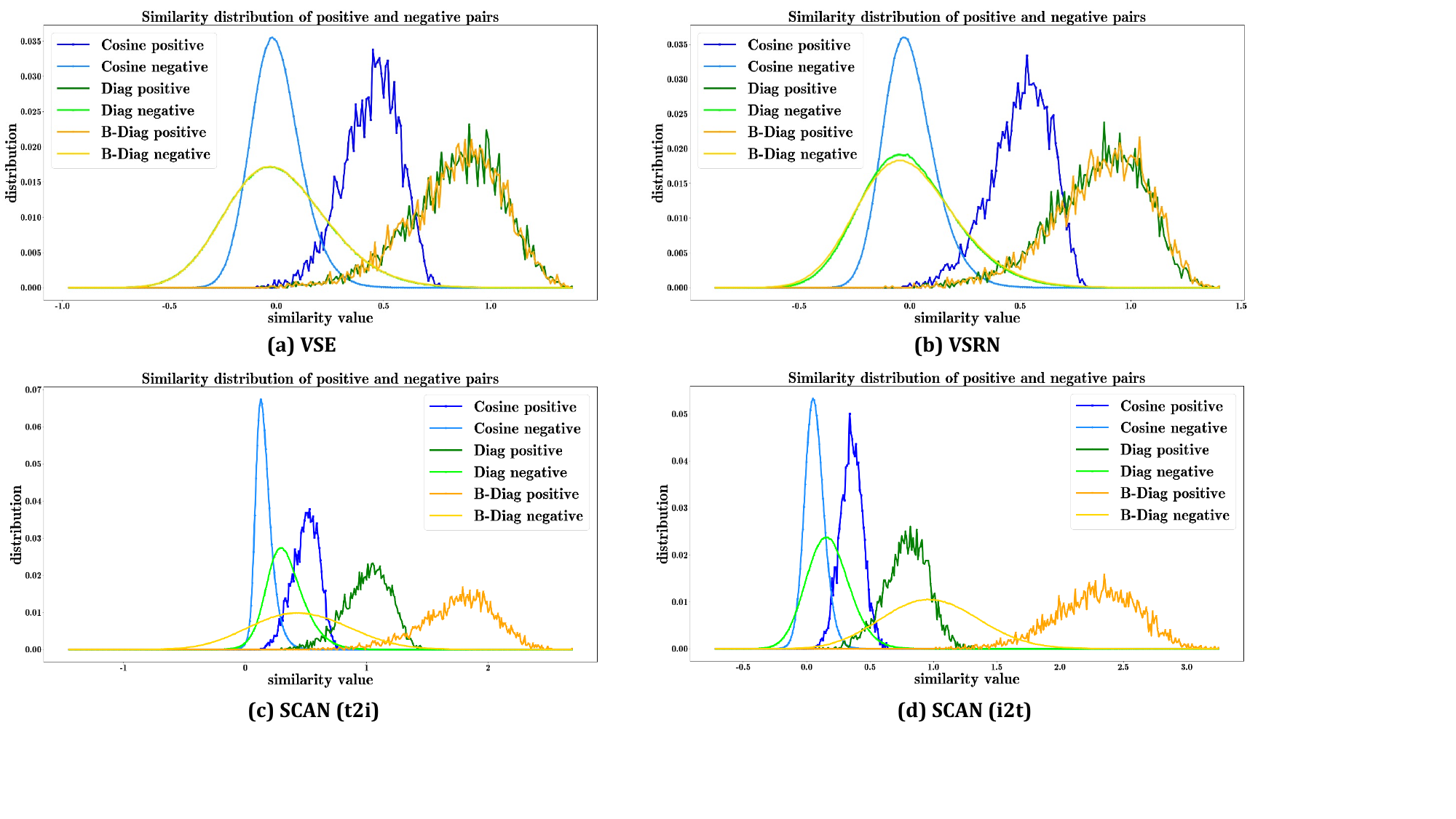} \\
    \end{tabular}
    \caption{Visualization of the learned similarity distributions of all positive and negative image-text pairs with Cosine/Diag/B-Diag metric on Flickr30K.}
    \label{fig:exp-similarity_distribution}
\end{figure*}

\begin{figure*}[t]
    \centering
    \begin{tabular}{@{}c}
    \includegraphics[width=0.95\linewidth,trim= 0 225 8 0,clip]{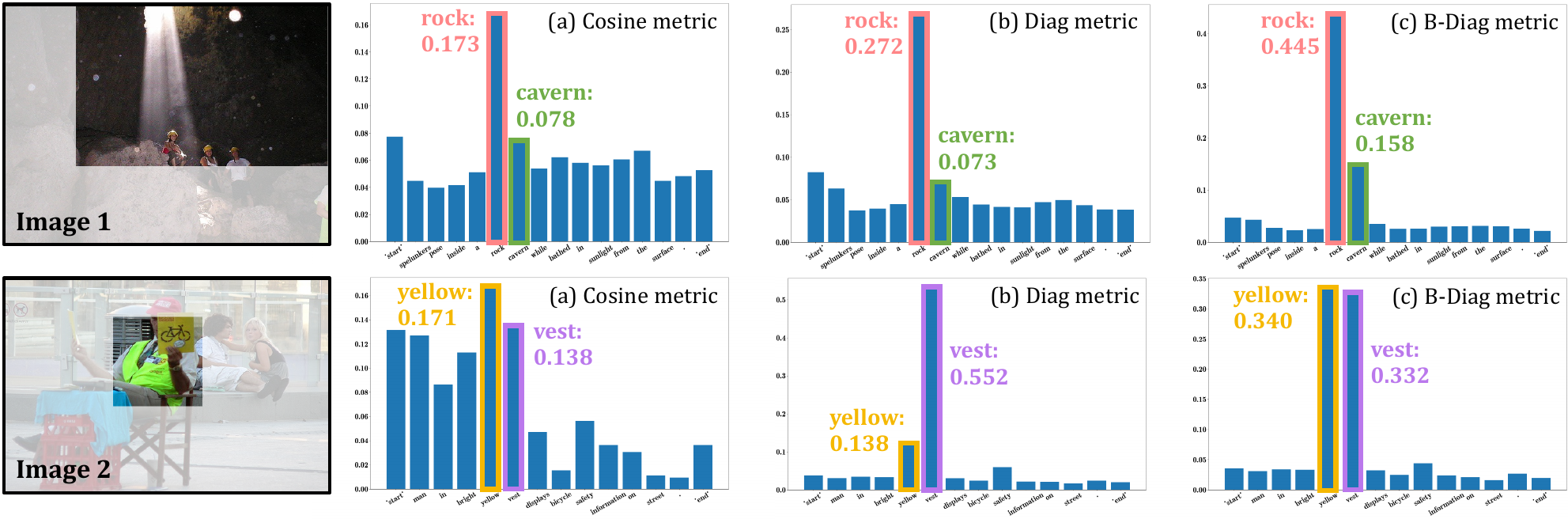} 
    \end{tabular}
    \caption{Visualization of image-to-text attention via SCAN (i2t). We compute the distributions over all words with Cosine/Diag/B-Diag metric on Flickr30K.}
    \label{fig:exp-image2text}
\end{figure*}
    
\subsection{Ablation Study}
\label{subsec:ablation-study}

\textbf{Influence of Cosine Initialization.} 
As shown in TABLE~\ref{tab:initializtion}, we report the impact of $\boldsymbol{W}$ initialization using identity matrix $\boldsymbol{I}$ or not based on the ensemble SCAN~\cite{ITM:SCAN} on Flickr30K. 
Comparing the Diag, B-Diag, and Dense metrics with their related versions without cosine-like initialization, we discover that such an initialization serves as a lubricant to accelerate model optimization and achieve performance improvements rapidly by imitating cosine-based channel connections at the beginning. Besides, this setting also applies to the dense distance metric. Notably, it not only speeds up convergence, but also consistently enhances the final performance gains. This is because optimizing a randomly-initialized matrix $\boldsymbol{W}$ disturbs the instinct within-channel correspondence, leading to optimization challenges and the risk of sub-optimal issues.
From the above observations, we initialize the $\boldsymbol{W}$ with an identity matrix $\boldsymbol{I}$ that would be updated during training.

\textbf{Configuration of Block Size.}
Fig.~\ref{fig:exp-block_size} shows the influence of varying block size $d$ on Flickr30K. 
Considering different feature dimensions~\cite{ITM:VSE++,ITM:VSRN,ITM:SCAN,ITM:BFAN}, we adjust the ratio $N$ ($D/d$) between the feature dimension $D$ and block size $d$ for universal configurations.
The best block sizes of VSRN~\cite{ITM:VSRN} and SCAN~\cite{ITM:SCAN} are within low-value and high-value scopes, respectively. Similar phenomena are also shown for other dual-encoder~\cite{ITM:VSE++,ITM:ESA,ReID:FastReID} and local-interaction~\cite{ITM:BFAN,ITM:IMRAM,ITM:NAAF} approaches.
Through Fig.~\ref{fig:exp-block_weight} and literature~\cite{CP:CW-NNK, CP:SRR-GR}, holistic representations tend to exhibit lower neighbor overlap and channel redundancy than local instance features. Consequently, the B-Diag variant excels at capturing fine-grained interactions, ideal for features with nuanced between-channel relationships, while the Diag variant is better suited for global features with its sparse correspondence and less channel dependency, offering a simpler yet more efficient solution.
Hence, we recommend the ratio $N$ ($D/d$) as 1024 for VSRN-like dual encoders, and 4 for SCAN-like local-interaction methods respectively, which shows the robustness and consistent improvements in TABLE~\ref{tab:cocof30k} and~\ref{tab:msmt_duke_sop}.

\begin{figure*}[t]
    \centering
    \begin{tabular}{@{}c}
    \includegraphics[width=0.96\linewidth,trim= 0 280 72 0,clip]{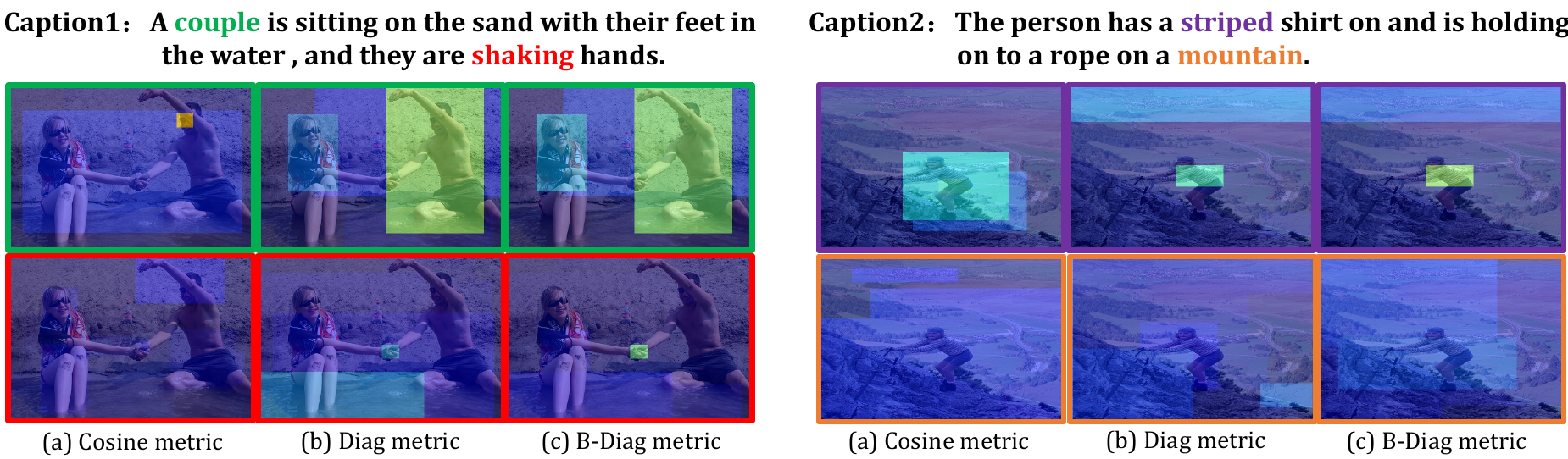}
    \end{tabular}
    \caption{Visualization of text-to-image attention via SCAN (t2i). We compute the distributions over all regions with Cosine/Diag/B-Diag metric on Flickr30K.}
    \label{fig:exp-text2image}
\end{figure*}

\begin{figure*}[t]
    \centering
    \begin{tabular}{@{}c}
    \includegraphics[width=0.96\linewidth,trim= 0 320 220 0,clip]{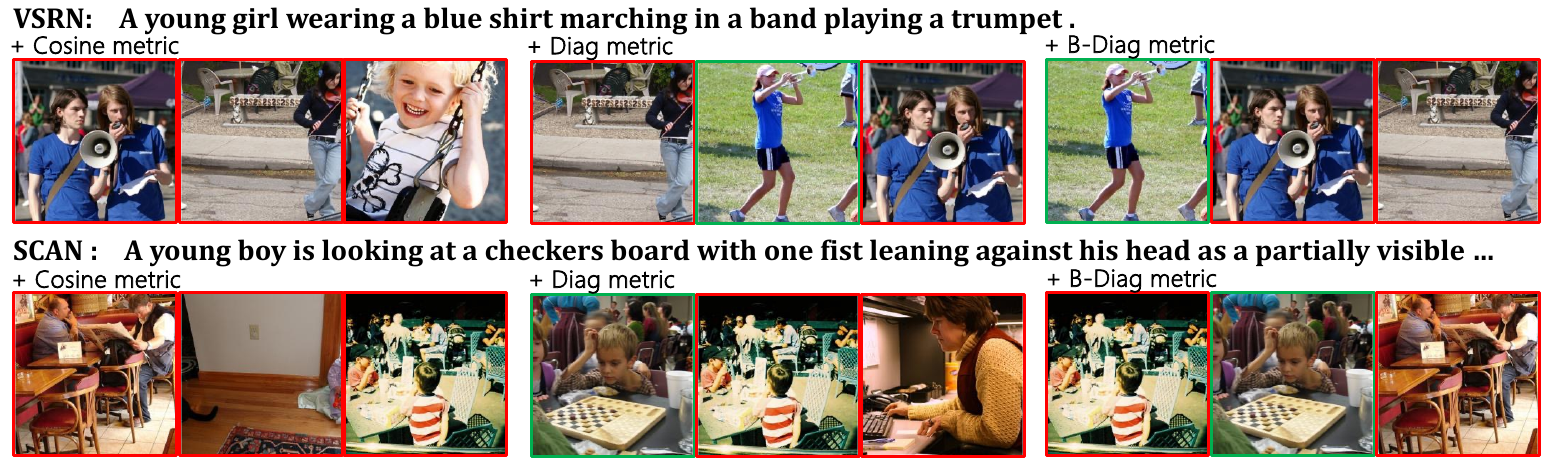}
    \end{tabular}
    \caption{Visualization of image retrieval based on VSRN and SCAN. Green border denotes the matched image and Red border denotes the unmatched image.}
    \label{fig:exp-image_retrieval}
\end{figure*}

\begin{figure*}[t]
    \centering
    \begin{tabular}{@{}c}
    \includegraphics[width=0.96\linewidth,trim= 0 275 90 0,clip]{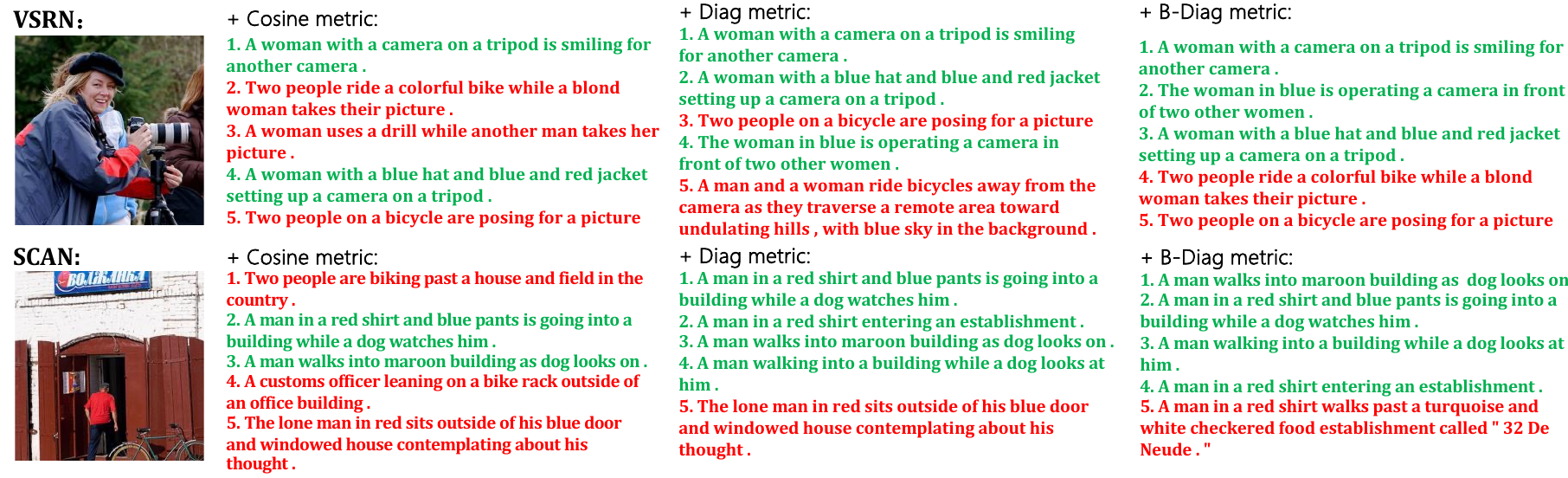}
    \end{tabular}
    \caption{Visualization of sentence retrieval based on VSRN and SCAN. Green font denotes the matched caption and Red font denotes the unmatched caption.}
    \label{fig:exp-sentence_retrieval}
\end{figure*}

\textbf{Visualization of Block Weight.}
To better understand the matching pattern across feature channels, Fig.~\ref{fig:exp-block_weight} illustrates the weight distributions by the B-Diag metric on Flickr30K. For better visualization, we reduce the size of matrix $\boldsymbol{W}$ to 64x64 through interval sampling. 
Here, the red and blue parts represent the positive and negative connection between channels respectively.
Overall, the weight coefficients, which signify positive, negative, and unrelated conditions across channels, range from approximately -0.5 to 2.4. Specifically, all within-channel weights are positive yet discrepant, diverging from the full-one connections of the cosine metric. 
Meanwhile, this diverse weight distribution assigns varying importance to different channels, enabling greater flexibility to adapt dynamically to optimal strategies across networks and datasets.
Besides, the diagonal elements occupy the maximum proportion and the final learned metric essentially enhances the cosine metric by incorporating more connections across channels while preserving sparsity in the connection matrix. This refinement significantly contributes to the powerful metric.

\textbf{Visualization of Similarity Distribution.} 
Fig.~\ref{fig:exp-similarity_distribution} displays the similarity values learned by each metric on Flickr30K. Here, the blue, green, and orange curves represent the positive and negative similarities of the Cosine, Diag, and B-Diag metrics, respectively. Unlike the Cosine metric, our methods extend the range of similarity scores while smoothing distance distributions by adjusting within-channel weights. Additionally, incorporating cross-channel weights allows for a more comprehensive measurement of cross-modal distances, further widening the gap between positive and negative samples.
It is worth noting that the distributions learned by the Diag and B-Diag metrics closely resemble each other in embedding-based methods. 
Our experiments reveal that in dual-encoder methods, the network tends to learn non-diagonal weights of the matrix that approach zero. This indicates that for holistic representations, the cross-channel correlation is minimal, resulting in the B-Diag metric collapsing into the Diag metric over time.
This phenomenon elucidates the necessity of introducing matrix sparsity and sheds light on why the latter occasionally shows marginal superiority over the former.

\textbf{Visualization of Cross-modal Attention.} 
Fig.~\ref{fig:exp-image2text} exhibits the image-to-text attention weights based on SCAN (i2t)~\cite{ITM:SCAN} on Flickr30K. We discover that our metrics effectively highlight correlations between regions and words while reducing interference from irrelevant content. Taking \textit{Image 2} as an example, the weights between image regions and "yellow vest" words computed by the Cosine metric are heavily influenced by other unrelated words. However, the Diag metric significantly increases their attention weights, ensuring clearer and more fine-grained correspondence across modalities. Furthermore, the B-Diag metric further enhances the importance of attribute-related word "yellow" due to the comprehensive matching projection facilitated by cross-channel connections. The similar phenomenon also occurs in the text-to-image attention weights depicted in Fig.~\ref{fig:exp-text2image} using SCAN (t2i)~\cite{ITM:SCAN}. 

\textbf{Visualization of Bi-directional Retrieval.} 
Fig.~\ref{fig:exp-image_retrieval} and Fig.~\ref{fig:exp-sentence_retrieval} display top retrieval results on image retrieval and sentence retrieval respectively. The top retrieved candidates for each query rank from left to right or top to bottom. The ground-truth and unmatched retrieval samples are shown in green and red marks respectively. In comparison with their raw metrics, our method enhances them to accurately discern positive samples, even in the presence of interference from hard negative samples. Additionally, the retrieved negatives exhibit a strong correlation and often contain similar semantics.

\section{Conclusion}
\label{sec:conclusion}
In this paper, we propose an innovative Generalized Structural Sparse Function to seek effective and efficient distance measurements between high-dimensional representation pairs. Specifically, it learns to systematically construct fine-grained interactions between channels and automatically adapt to the powerful similarity strategy across modalities. Extensive experiments on six diverse cross-modal and uni-modal benchmarks validate that our metric significantly facilitates retrieval performance over various popular matching approaches, enjoying both portability and efficiency. We further demonstrate its great potential and broad applicability on a variety of promising scenarios, \eg, Attention Mechanism, Token-wise Alignment, Knowledge Distillation, and Transfer Learning.

\textbf{Discussion.} We believe this work serves as an important inspiration for a desired, powerful, and more importantly, efficient distance metric in deep cross-modal metric learning. Such structural metrics are compatible with rich application scenarios aiding flexibility and efficiency, but a key issue is the empirical size and shape of block partition that deserves further research. In the future, we aim to train the connection matrix $\boldsymbol{W}$ and adjacent strategy $\boldsymbol{U}$ simultaneously that can be both updated in an end-to-end manner by backward gradients.

\ifCLASSOPTIONcaptionsoff
  \newpage
\fi
\bibliographystyle{IEEEtran}
\bibliography{IEEEabrv,refs}
\end{document}